\documentclass[letterpaper, 10 pt, conference]{ieeeconf}

\IEEEoverridecommandlockouts
\usepackage{times}
\usepackage{mathptmx}
\usepackage{amsmath,amssymb,mathtools}
\usepackage{graphicx}
\usepackage{algorithm}
\usepackage{algpseudocode}
\usepackage{booktabs,multirow,tabularx,threeparttable}
\usepackage{caption}
\usepackage{stfloats}
\usepackage{placeins}
\usepackage{url}
\usepackage[noadjust]{cite}

\usepackage{xcolor}
\usepackage{fancyhdr}

\newcommand{\imgscale}{0.8}

\fancypagestyle{arxivnotice}{
    \fancyhf{}%
    \fancyhead[C]{%
        \parbox{\textwidth}{%
            \centering
            \large\color{gray}
            This paper has been accepted for publication at the\\
            IEEE/RSJ International Conference on Intelligent Robots and
            Systems (IROS), 2026.\ \copyright\ IEEE
        }%
    }%
}

\title{\LARGE \bf
AeroGrab: A Unified Framework for Aerial Grasping in Cluttered Environments
}

\author{
Shivansh Pratap Singh$^{1*}$,
Naveen Nair$^{1*}$,
Samaksh Ujjwal$^{1}$,
Sarthak Mishra$^{1}$,
Soham Patil$^{1}$,\\
Rishabh Dev Yadav$^{2}$,
Spandan Roy$^{1}$%
}

\begin{document}

\twocolumn[{
\renewcommand\twocolumn[1][]{#1}

\maketitle

\vspace{-4mm}

\begin{center}
    \centering
    \includegraphics[width=0.90\textwidth]{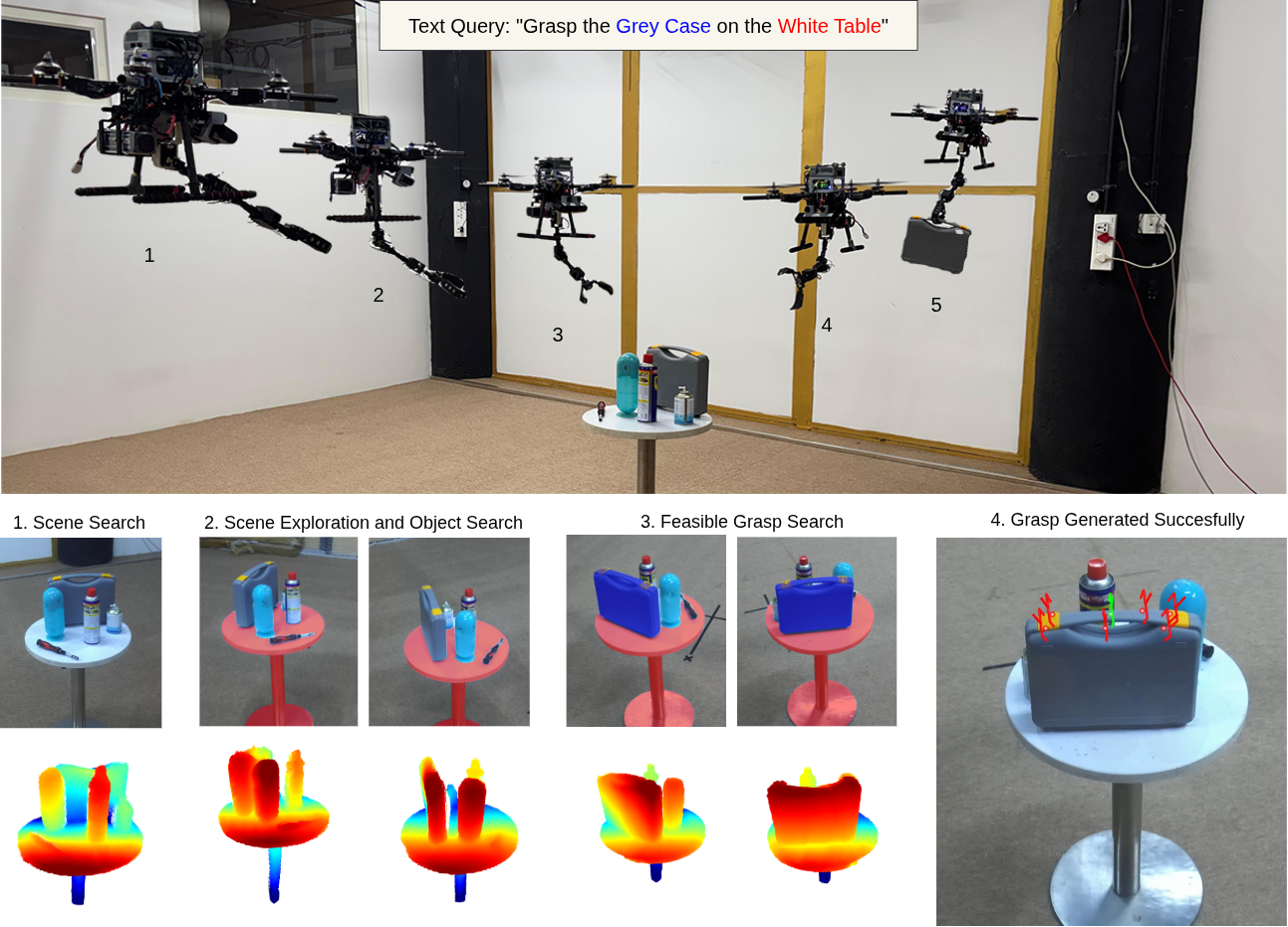}
    \captionof{figure}{
        Architecture Overview. A unified aerial grasping framework that
        tightly couples language-guided semantic perception with
        platform-aware kinematic and collision constraints, enabling safe
        interaction in a cluttered environment.
    }
    \label{fig:architecture}
\end{center}

\vspace{-2mm}
}]

\thispagestyle{arxivnotice}

\begingroup
\renewcommand{\thefootnote}{}
\footnotetext{
    This work is supported partly by the ``Edge-AI-GGCNN'' project from
    Qualcomm Technologies and partly by the ``UASAT'' project sponsored
    by MeitY, India. $(*)$ denotes equal contribution.
}
\endgroup

\begingroup
\renewcommand{\thefootnote}{\arabic{footnote}}

\footnotetext[1]{
    Robotics Research Center, IIIT Hyderabad, India.
    Emails:
    {\footnotesize\ttfamily
    \{shivansh.singh, samaksh.ujjawal, sarthak.mishra,
    soham.patil\}@research.iiit.ac.in},
    {\footnotesize\ttfamily naveennair2003@gmail.com},
    {\footnotesize\ttfamily spandan.roy@iiit.ac.in}.
}

\footnotetext[2]{
    Department of Computer Science, University of Manchester, UK.
    Email:
    {\footnotesize\ttfamily
    rishabh.yadav@postgrad.manchester.ac.uk}.
}

\endgroup

\begin{abstract}
Reliable aerial grasping in cluttered environments remains challenging due to occlusions and collision risks. Existing aerial manipulation pipelines largely rely on centroid-based grasping and lack integration between the grasp pose generation models, active exploration, and language-level task specification, leaving a gap in complete end-to-end operation. In this work, we present an integrated pipeline for reliable aerial grasping in cluttered environments. Given a scene and a language instruction, the system identifies the target object and actively explores it to gain better views of the object. During exploration, a grasp generation network predicts multiple 6-DoF grasp candidates for each view. Each candidate is evaluated using a collision-aware feasibility
framework, and the overall best grasp is selected and executed using standard trajectory generation and control methods. Experiments in cluttered real-world scenarios demonstrate robust and reliable grasp execution, highlighting the effectiveness of combining active perception with feasibility-aware grasp selection for aerial manipulation. 
\noindent\textbf{Video:} \url{https://youtu.be/aSCQm8HdasI}
\end{abstract}

\section{Introduction}

Aerial manipulators have gained significant attention for performing contact-rich interaction tasks such as grasping, picking, and relocating objects in environments that are difficult to access with ground robots \cite{Ruggiero2018LRA, Ollero2018RAM, yadav2025integrated, yadav2024modular}. Successfully executing such tasks requires the aerial manipulator to detect and locate the target object, estimate a feasible grasp pose, and plan actions that ensure safe interaction with the environment while considering the reachability of the manipulator and avoiding collisions in cluttered scenes. Many existing aerial manipulation systems simplify grasp pose generation by targeting the geometric center of the object \cite{bauer2025open, bauer2023autonomous}. While this strategy can be effective for homogeneous and compact objects, centroid-based approaches often fail for everyday objects whose stable grasp regions lie on functional components such as handles, rims, or edges. Furthermore, most aerial manipulation systems assume that the target object is clearly visible. In cluttered environments, however, objects are frequently partially occluded, making feasible grasp pose estimation highly viewpoint-dependent. Consequently, the aerial platform must actively reposition to improve object observability while maintaining safety around nearby obstacles and respecting platform constraints \cite{Bajcsy2018RevisitingAP}. Despite these challenges, most existing aerial manipulation pipelines \cite{Ubellacker2024npjRobotics, thomas2014toward} do not integrate active perception with collision-aware grasp pose generation, which limits their reliability when operating in cluttered environments.

To address these limitations, we propose a framework (Fig.~\ref{fig:architecture}) that enables language-guided 6-DoF feasible grasp generation for aerial manipulators operating in cluttered scenes. The proposed system integrates perception, grasp feasibility analysis, and collision-aware planning to allow the aerial manipulator to actively reposition, identify functional grasp regions, and execute safe grasping actions while avoiding collisions with surrounding obstacles. To ensure reliable execution, the system integrates a minimum-snap trajectory generator \cite{Mellinger2011MinimumSnap} together with an adaptive control \cite{yadav2025integrated} for robust aerial manipulation. The proposed framework is validated through extensive simulation and hardware experiments, demonstrating successful grasping and object retrieval across diverse objects in cluttered environments.

\section{Related Work and Contribution}

Early research in aerial manipulation focused on enabling stable physical interaction between uncrewed aerial vehicles and the environment through advances in modeling, control, and mechanical design \cite{orsag2018aerial}. Initial works demonstrated stable contact and whole-body control using rigid or compliant manipulators mounted on aerial platforms \cite{Kim2013IROS,Fumagalli2014RAM,Bartelds2016RAL}. Subsequent systems incorporated onboard perception and visual feedback to enable autonomous grasping and manipulation tasks \cite{Chen2019Sensors,Oberson2022IROS}. Surveys of aerial robots with grasping and perching capabilities are provided in \cite{meng2020survey,Meng2022Frontiers}. More recent works integrate perception, planning, and control to improve autonomy and robustness in aerial interaction, including vision-based grasping, force-aware grasp execution, and compliant contact control \cite{Ubellacker2024npjRobotics,chen2025aerial,Hoi2026arXiv,Zhan2026arXiv}.
To reduce reliance on exact analytical models, few studies have learned residual dynamics~\cite{das2025dronediffusion, ujjawal2026learn, yadav2026learning, ujjawal2025aermani, yadav2025arcade, yadav2026physics}. Few works also use language and vision models for placement, vision language model-based skill selection~\cite{mishra2026aeroplace, mishra2025aermani, song2025soranav}.

Despite these advances, aerial grasping in cluttered environments remains comparatively underexplored. Many systems \cite{ramon2020grasp,bauer2025open,Meng2022Frontiers} assume that the target object is clearly visible or that object segmentation is available beforehand. In realistic environments, however, objects are often partially occluded, making grasp feasibility strongly viewpoint-dependent and requiring the aerial platform to actively reposition to improve visibility before reliable grasp synthesis can be performed. 

Beyond core aerial manipulation, recent robotics research has turned its attention toward more intuitive task specification via natural language. Vision--language and large language models (VLMs/LLMs) now enable robots to ground complex instructions within visual scenes and isolate task-relevant objects \cite{Shridhar2020ALFRED,Lynch2023RT2,Ahn2022SayCan}. Concurrently, progress in safe motion generation has leveraged simplified robot representations and visual sensing. Techniques such as null--space saturation for redundant manipulators \cite{flacco2015control} and image-space planning using visual keypoints \cite{chatterjee2025image} have demonstrated significant efficacy in ground-based manipulation. However, the integration of these advances into aerial systems remains a largely unexplored research frontier.

Motivated by these limitations, we couple viewpoint adaptation with grasp synthesis and feasibility evaluation. Our system integrates language-guided target grounding, active viewpoint selection, and collision-aware 6-DoF grasp generation. Unlike approaches that reason only about the manipulator arm or rely on simplified visual abstractions, we project a convex-hull approximation of the full aerial manipulator body into the camera frame and evaluate potential collisions directly against sensed depth during grasp selection, enabling safe execution with lightweight onboard computation.

\noindent Our main contributions are as follows:

\begin{itemize}

\item \textbf{Language-guided aerial grasping in clutter:}  
We present a unified aerial manipulation pipeline that translates natural language instructions into executable grasping actions by combining language-guided target grounding, perception, and aerial manipulation.

\item \textbf{Active perception for viewpoint-dependent grasping:}  
We introduce an active viewpoint adaptation strategy that allows the aerial manipulator to autonomously reposition around cluttered scenes to improve target observability and expose feasible grasp approaches.

\item \textbf{Collision-aware 6-DoF grasp feasibility evaluation:}  
We propose a feasibility ranking module that evaluates candidate grasp poses by explicitly considering full-body approach clearance, platform reachability, and surrounding clutter, enabling safe grasp selection simultaneously with active collision checking.

\end{itemize}

\section{Methodology}

\begin{figure*}[t]
    \centering
    \includegraphics[width=0.95\textwidth]{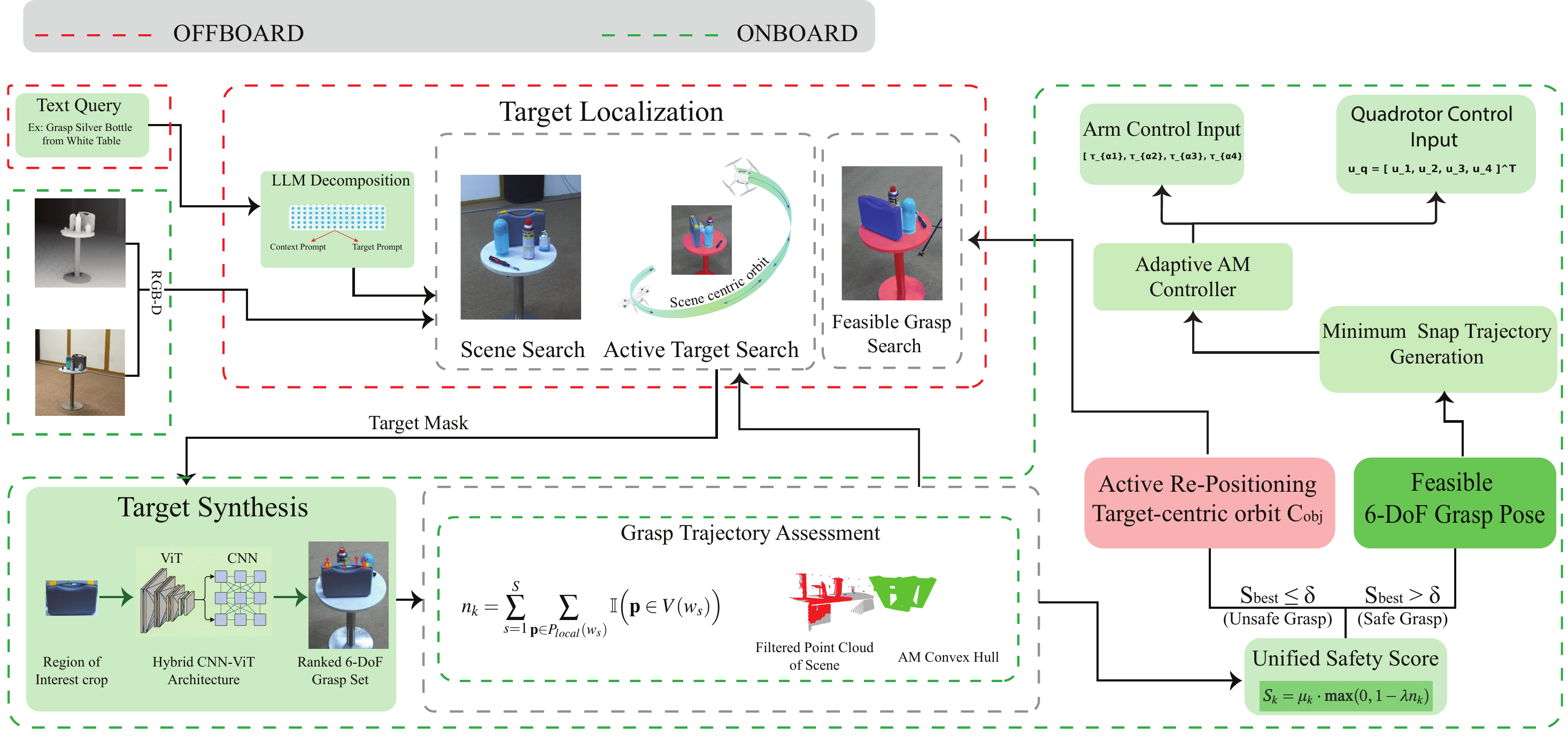}
    \caption{Unified perception, planning, and control pipeline for aerial grasping in clutter. The framework connects language grounding, active perception, 6-DoF grasp generation, platform-aware feasibility analysis, and collision-safe execution.}
    \label{fig:Pipeline}
\end{figure*}

This section describes the architecture and implementation of the proposed aerial grasping framework. The objective is to convert language instructions into collision-safe aerial grasps through integrated planning, control, and streamed segmentation. An overview of the complete pipeline is illustrated in Fig.~\ref{fig:Pipeline}.

\subsection{Language-Guided Target Localization}
The pipeline begins by converting the natural language instruction into structured visual targets. 
Given a command such as ``Grasp the green bottle from the steel table,'' the system first decomposes the instruction using an LLM into two structured prompts: a scene reference prompt $T_{ctx}$ (``steel table'') and a target reference prompt $T_{obj}$ (``green bottle''). The platform first localizes the visually dominant scene before identifying the graspable object.

The aerial platform initially performs a stationary yaw-based scan to search for $T_{ctx}$ within the camera field of view. During this phase, we employ SAM-3~\cite{Carion2025SAM3} for real-time segmentation due to its strong zero-shot generalization and ability to produce stable object masks from sparse prompts. Once the scene is detected, the model maintains the segmentation mask consistently across subsequent frames, providing a reliable visual reference for downstream perception and grasp planning.

Using the segmented region and its associated depth measurements, the system estimates the 3D centroid $\mathbf{c}_{ctx}$ of the scene surface. The drone then transitions from stationary scanning to an active orbital exploration motion around $\mathbf{c}_{ctx}$, inspired by the active perception strategies that seek viewpoint diversity to reveal occluded objects \cite{Bajcsy2018RevisitingAP}. During this motion, the orbit radius is regulated so that the entire scene volume remains within the camera's field of view (FoV). The desired orbital radius $r_{goal}$ is adapted based on the spatial extent of the segmented scene region:
\begin{equation}
    r_{goal} = \frac{\max(\|\mathbf{p} - \mathbf{c}_{ctx}\|)}
    {\tan(\text{FoV}/2)}, \quad \forall \mathbf{p} \in {P}_{ctx}
\end{equation}
where ${P}_{ctx}$ denotes the set of 3D points belonging to the segmented scene surface, and $\mathbf{p}$ represents an individual 3D point in this set expressed in the world frame.

The velocity command is defined as
\begin{equation} \label{eq:orbit_velocity}
    \mathbf{v}_{cmd} =
    v_{orb} \cdot \mathbf{t}_{tan} +
    k_p \left(r_{goal} - \|\mathbf{p}_{B} - \mathbf{c}_{ctx}\|\right)
    \mathbf{n}_{rad},
\end{equation}
where $\mathbf{v}_{cmd}$ denotes the commanded translational velocity of the aerial platform, $v_{orb}$ is the desired tangential orbiting speed, $k_p$ is a proportional gain regulating the radial distance, $\mathbf{p}_{B}$ represents the current position of the drone in the world frame, $\mathbf{t}_{tan}$ is the unit tangential direction along the orbit, and $\mathbf{n}_{rad}$ is the unit radial direction pointing toward the orbit center.

As the drone orbits, SAM-3 continues to update the segmentation masks, allowing the system to detect the target object $T_{obj}$ once it enters the visible region. When the target mask $M_{obj}$ becomes available, the system has both a localized object region and its associated depth measurements. At this stage, the problem shifts from scene exploration to selecting a physically executable grasp.

\subsection{Target Synthesis}
\label{sec:target_synthesis}
Given the segmented target region, we extract a cropped Region of Interest (RoI) $I_{crop}$ from the RGB-D stream centered on $M_{obj}$. This crop isolates the object from the cluttered scene while preserving its local geometric structure.

The objective of this stage is to generate a diverse set of candidate 6-DoF grasps directly from the RGB-D observation. Each grasp is parameterized by a translation \textbf{t}, orientation \textbf{R}, gripper width \textbf{w}, and an associated confidence score \textbf{s}.

To generate these candidates, we use a Hybrid CNN-ViT architecture trained on the GraspNet-1Billion \cite{Fang2020GraspNet1Billion} dataset. Given the cropped RGB-D input, the network predicts a fixed set of 6-DoF grasp poses and associated quality scores, forming the raw candidate set ${G}_{raw}$.

Since GraspNet annotations are designed primarily for fixed-base manipulators, many predicted grasps are infeasible for aerial platforms. To address this mismatch, we filter the raw grasp set using aerial feasibility constraints that reflect the physical limitations of aerial manipulation. Specifically, we enforce the following constraints:

\begin{itemize}
\item \textbf{Gravity-aligned approach.} Grasps whose approach direction opposes gravity are discarded, as aerial robots typically access objects from above and cannot execute grasps requiring access beneath support surfaces.

\item \textbf{Stability-aware scoring.} Grasps located far from the object centroid introduce larger moment arms during lift-off that can destabilize the vehicle. Instead of removing these grasps entirely, their scores are penalized according to their centroid distance so that they remain available in cluttered scenes where collision constraints may eliminate more stable alternatives.
\end{itemize}

The final grasp score is computed as
\begin{equation}
s_{final} = s \cdot \exp\left(-\alpha \| \mathbf{t} - c_{obj} \| \right)
\end{equation}

where $s$ denotes the network confidence, $c_{obj}$ is the 3D centroid and $\alpha$ controls the strength of the stability penalty. After applying these constraints and re-weighting, the remaining grasps form the feasible set $G_{feasible}$, which contains multiple physically executable grasp candidates suitable for aerial manipulation.

\subsection{Collision Feasibility Filtering} \label{sec:collision_filtering}

Although visually valid, predicted grasps may be infeasible for an aerial platform. Each candidate is therefore filtered using kinematic reachability and collision constraints prior to execution.

The grasp network predicts candidate poses in the camera frame, which are transformed into the world frame using the onboard visual--inertial state estimate. For each candidate, inverse kinematics is solved to determine whether the manipulator can reach the corresponding end-effector pose without violating joint limits. Grasps with no valid joint solution are discarded.

Even when a grasp is kinematically reachable, executing it may still be unsafe due to potential collisions between the manipulator, drone body, and surrounding obstacles during the approach motion. 

\subsubsection{Batched Trajectory Collision Evaluation}

Once we isolate the set of kinematically reachable grasps, we must verify that the drone can physically fly to them without colliding with the environment. Checking each grasp sequentially would be too slow for a real-time control loop. Instead, we formulate a batched process that evaluates the approach paths for all candidate grasps simultaneously.

Let the visual output be defined over an image grid of size $W \times H$. For the $k$-th valid pixel $(u, v)$ belonging to the target object, the synthesis network provides a grasp candidate $g_k$ with an initial visual quality score $v_k$. 

For each grasp $g_k$, we define the required target state using the spatial pose of the drone base $T_k \in SE(3)$ and the corresponding manipulator joint configuration $\mathbf{q}_k \in \mathbb{R}^4$ derived from the inverse kinematics step. We then generate a full-body approach trajectory $\tau_k$ by linearly interpolating between the robot's current state $(T_{curr}, \mathbf{q}_{curr})$ and the target state $(T_k, \mathbf{q}_k)$. Mathematically, we discretize this trajectory into a set of $S$ waypoints:
\begin{equation}
    \tau_k = \{ w_1, w_2, \dots, w_S \}
\end{equation}

To evaluate environmental collisions along these waypoints, we approximate the robot's physical volume as a set of convex hulls, denoted as $H_{robot}$. By representing these convex hulls algebraically as a set of planar half-spaces (defined by face normals and offsets), we can convert the 3D intersection problem into highly parallelized matrix multiplications. This specific formulation allows the onboard GPU to evaluate thousands of scene points across hundreds of trajectory waypoints in real time.

We execute this collision check in two sequential steps to maximize computational efficiency:
\begin{enumerate}
    \item \textbf{Coarse Filtering:} First, we perform a rapid spatial query using a KD-tree representation of the environment's point cloud $P_{scene}$. We project a conservative bounding sphere around each trajectory waypoint $w_s$. If a sphere contains zero environmental points, that specific segment of the path is guaranteed to be safe, bypassing further calculations.
    \item \textbf{Exact GPU Intersection:} If a bounding sphere contains points, we extract that localized subset $P_{local}(w_s)$ and pass it to the GPU. The GPU evaluates whether these points fall strictly inside the geometric boundaries of the robot's convex hulls at that specific waypoint.
\end{enumerate}

Let $V(w_s)$ represent the spatial volume occupied by the robot hulls at waypoint $w_s$. For a given grasp candidate $g_k$, the total collision penalty $n_k$ is calculated by aggregating every environmental point $\mathbf{p}$ that penetrates the robot's volume across all $S$ waypoints of the approach path:
\begin{equation} \label{eq:collision_penalty}
    n_k = \sum_{s=1}^{S} \sum_{\mathbf{p} \in P_{local}(w_s)} \mathbb{I} \Big( \mathbf{p} \in V(w_s) \Big)
\end{equation}
where the indicator function $\mathbb{I}(\cdot)$ returns 1 if the point $\mathbf{p}$ lies inside the robot volume, and 0 otherwise. By summing across all discrete waypoints, this formulation can be used to calculate a collision penalty across the duration and depth of the trajectories. Any trajectories that dwell inside obstacles accumulate an exponentially higher penalty than those resulting in brief, grazing contacts.

We use this collision penalty to adjust the network's original visual score $v_k$. The final execution score $S_k$ is computed as:
\begin{equation} \label{eq:execution_score}
    S_k = v_k \cdot \max(0, 1 - \lambda n_k)
\end{equation}
where $\lambda$ is a tunable penalty scalar. This equation acts as a soft constraint. It heavily penalizes grasps that require the drone to fly directly through dense obstacles, driving their score to zero. At the same time, it is forgiving enough to allow grasps that might involve minor, grazing contacts with the environment, which are often unavoidable when interacting with dense clutter.

\subsection{Decision Logic and Grasp Execution} \label{sec:decision_logic}
The final phase of the pipeline arbitrates the physical execution based on the highest safety-weighted score, $S_{best}$.

\subsubsection{Active Re-Positioning (Unsafe Grasp)}
If all candidates are deemed kinematically unreachable or severely obstructed ($S_{best} \leq \delta$), where $\delta$ is a predefined feasibility threshold, the system declines to execute the grasp. Instead of failing, the drone autonomously continues the active orbiting behavior defined in Eq. (2). However, the guidance law is now locked onto the specific \textit{Target Centroid} $\mathbf{c}_{obj}$ rather than the broader scene. This target-centric orbit continuously updates the environmental point cloud, systematically exploring the kinematic manifold until a collision-free geometric corridor is revealed.

\subsubsection{Approach and Execution (Safe Grasp)}
Once a valid candidate ($S_{best} > \delta$) is identified, the system transitions to the execution phase. The manipulator gripper is opened and the arm joints are pre-positioned to $\mathbf{q}_{sol}$. The UAV then follows the planned approach trajectory toward the grasp pose $\tau(g^*)$. Upon reaching the pre-grasp waypoint, the drone holds a short stabilizing hover to dissipate aerodynamic disturbances. The gripper is then actuated to secure the object, followed by a vertical translation phase to clear the surrounding clutter.

\section{Control and Trajectory Execution}
Once a valid grasp is selected, the aerial manipulator executes a dynamically feasible approach trajectory generated using a minimum-snap trajectory \cite{Mellinger2011MinimumSnap} formulation based on differential flatness. The generated trajectory is tracked using an adaptive controller \cite{yadav2025integrated}, which compensates for modelling uncertainties and external disturbances during close-proximity interaction. All experiments use identical trajectory and control parameters.

\begin{algorithm}[t]
\caption{Language-Guided Aerial Grasping}
\label{alg:main_pipeline}
\footnotesize
\begin{algorithmic}[1]
\Require Command $\mathcal{L}_{cmd}$ and RGB-D stream $I_t$
\State $(T_{ctx},T_{obj})\gets\mathrm{LLM}(\mathcal{L}_{cmd})$

\Loop
    \State $(M_{ctx},M_{obj})\gets
    \mathrm{SAM\text{-}3}(I_t,T_{ctx},T_{obj})$

    \If{$M_{ctx}=\emptyset$}
        \State \Call{YawSearch}{}

    \ElsIf{$M_{obj}=\emptyset$}
        \State \Call{Orbit}{$M_{ctx}$}
        \Comment{Eq.~\eqref{eq:orbit_velocity}}

    \Else
        \State $\mathcal{G}\gets
        \Call{GenerateGrasps}{I_t,M_{obj}}$
        \Comment{Sec.~\ref{sec:target_synthesis}}

        \State $(g^*,S^*)\gets
        \Call{RankFeasibleGrasps}{\mathcal{G},P_{scene}}$
        \Comment{Eqs.~\eqref{eq:collision_penalty}--\eqref{eq:execution_score}}

        \If{$S^*>\delta$}
            \State \Call{ExecuteAndLift}{$g^*$}
            \State \Return \textsc{Success}
        \Else
            \State \Call{Orbit}{$M_{obj}$}
            \Comment{Sec.~\ref{sec:decision_logic}}
        \EndIf
    \EndIf
\EndLoop
\end{algorithmic}
\end{algorithm}

\section{Experiments}

\begin{figure*}[t]
    \centering
    \includegraphics[width=0.85\textwidth]{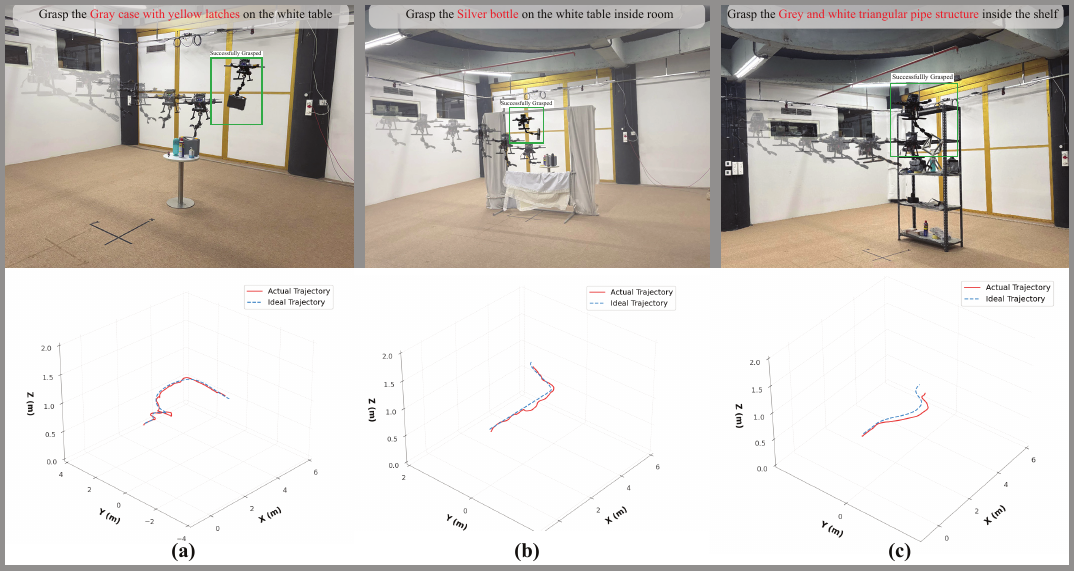}
    \caption{Experiment Scenarios. (a) Tabletop clutter, (b) window-constrained access, (c) shelf reachability test. For each scenario, we show the scene view, trajectory, and successful grasp.}
    \label{fig:experiments}
\end{figure*}

We evaluate the proposed framework through a series of simulated and real-world experiments designed to test its reliability in unstructured environments. Specifically, our evaluation focuses on three primary objectives: \textit{(i)} the accuracy of the Hybrid CNN-ViT grasp synthesis in dense tabletop clutter, \textit{(ii)} the robustness of the semantic tracking pipeline under continuous platform ego-motion, and \textit{(iii)} the effectiveness of the active collision-informed search in ensuring kinematically feasible and collision-free approach trajectories.

\begin{figure}[t]
    \centering
\includegraphics[width=\imgscale\linewidth]{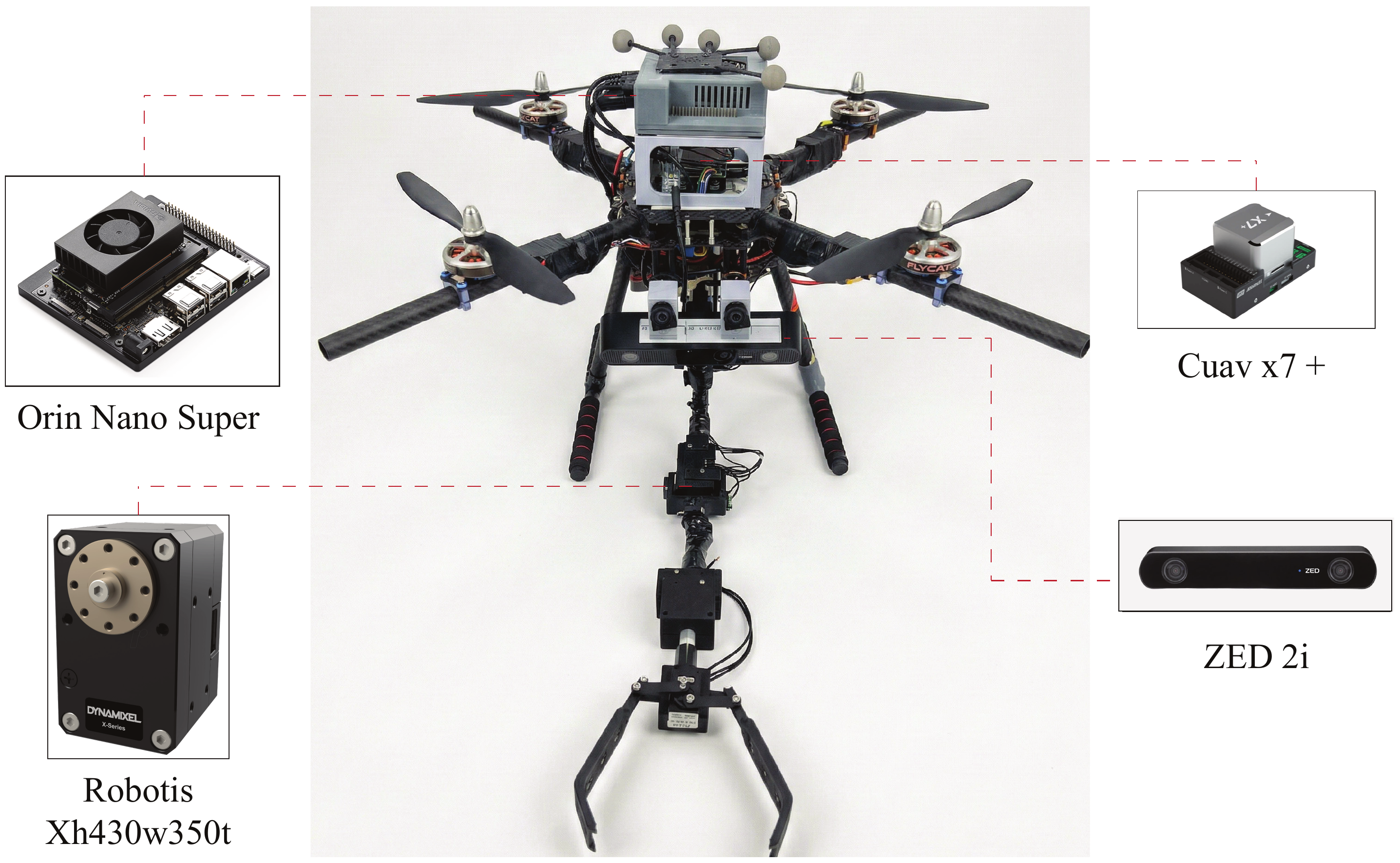}
    \caption{\small System Hardware Overview.}
    \label{fig:hardware}
\end{figure}

\subsection{Experimental Platform}
Our aerial manipulation platform uses a custom 450\,mm quadrotor frame (Fig.~\ref{fig:hardware}) with a 2.5\,kg takeoff weight excluding payload. It is designed for real-time, perception-driven aerial grasping.
The vehicle is powered by FlyCat 5010 750\,KV BLDC motors with 12-inch propellers and controlled by a CUAV X7+ flight controller running PX4. PX4 performs low-level attitude stabilization, while high-level commands are exchanged through ROS2 Humble via the PX4 DDS interface.
An NVIDIA Orin Nano Super onboard computer executes grasp synthesis, trajectory generation, and control, using visual input from a ZED~2i stereo RGB-D camera. The manipulator is a lightweight 4-DoF serial arm driven by Dynamixel XH430-W350 servos and powered through a regulated 12\,V buck converter.
An offboard workstation with an NVIDIA RTX~4090 GPU and AMD Ryzen~9 processor communicates through ROS2 DDS and handles visualization and target segmentation. All time-critical perception, planning, and control processes run onboard.
Experiments are conducted in an OptiTrack motion-capture environment, which provides high-precision pose measurements for evaluation and controller validation during aerial grasping.

\begin{figure}[t]
    \centering
    \includegraphics[width=0.25\textwidth]{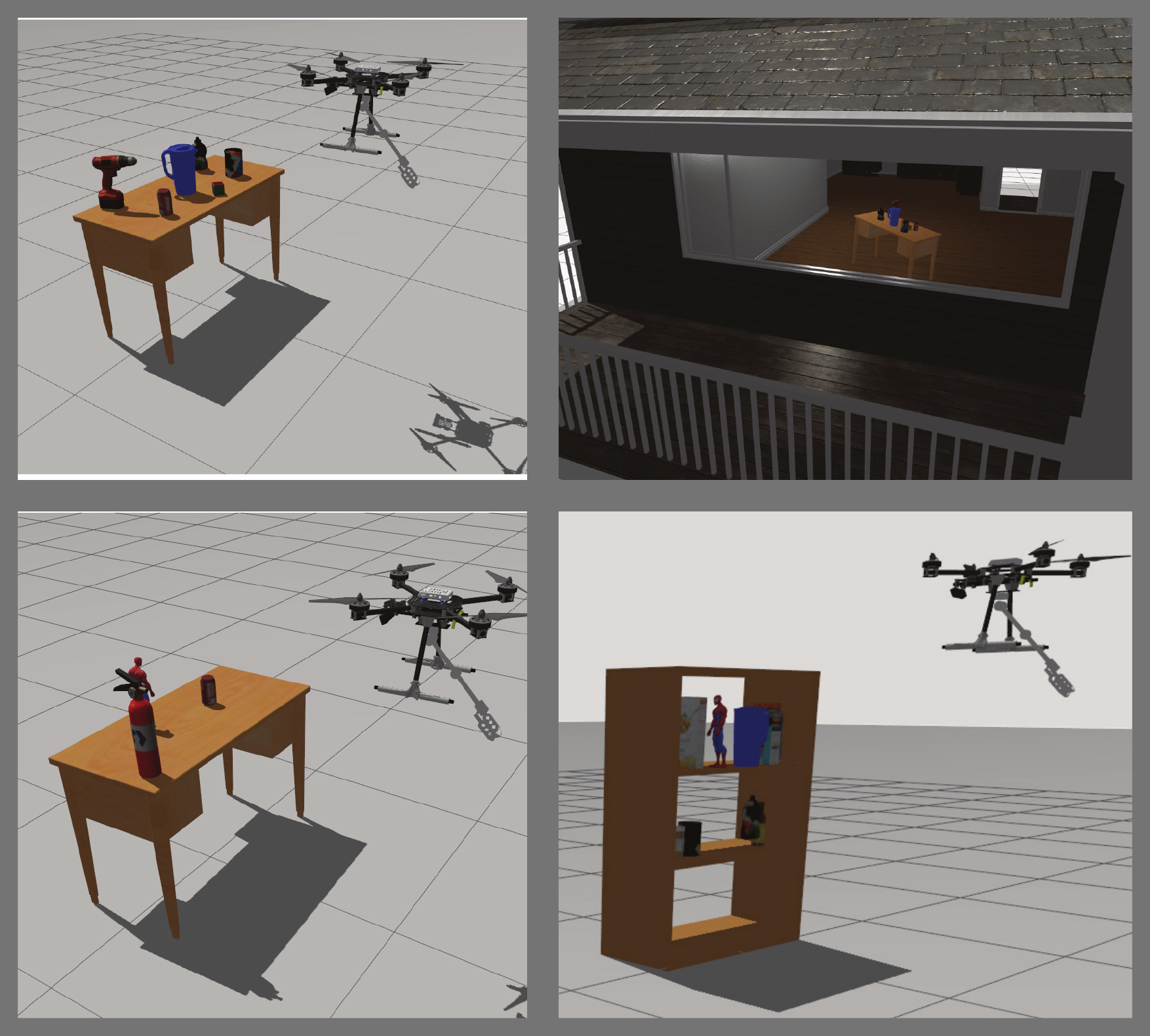}
    \caption{{Simulation Setup Overview.}}
    \label{fig:sim}
\end{figure}

\subsection{Simulation Environment and Setup}
We validate the proposed active search and collision-aware grasping pipeline in a high-fidelity Gazebo Harmonic simulation prior to hardware deployment. Quadrotor dynamics and low-level attitude control are simulated using the PX4 Software-In-The-Loop (SITL) stack. A 4-DoF (4R) manipulator is simulated via \texttt{ros2\_control} in torque-control mode to capture arm--base dynamic coupling during approach and grasp execution.

We use an OAK-Lite RGB-D sensor model in simulation and inject Gaussian depth noise ($\sigma=0.02$\,m) with depth-dropout artifacts to reduce sim-to-real mismatch. (Fig.~\ref{fig:sim})

\subsection{Scenario Definitions and Task Procedures}
We evaluated three task scenarios, each targeting a distinct failure mode of aerial manipulation. In each trial, the AM receives a natural-language query specifying the target object, performs an active visual search to localize it, selects a feasible 6-DoF grasp, and then approaches and executes the grasp while maintaining target association under viewpoint changes and occlusions.

\smallskip

\smallskip
\noindent\textbf{Scenario 1 (Tabletop Clutter):}
Dense tabletop clutter with partial overlaps; evaluate grasp selection and collision-feasible approach in near-field clutter. 
(Fig.~\ref{fig:experiments}(a), Table~\ref{tab:grasp_results})

\smallskip
\noindent\textbf{Scenario 2 (Window / Aperture-Constrained):}
Target must be approached through a narrow window, enforcing strict base-clearance and partial observability. 
(Fig.~\ref{fig:experiments}(b), Table~\ref{tab:grasp_results})

\smallskip
\noindent\textbf{Scenario 3 (Shelf / Reachability \& Stability):}
Multi-tier shelf with restricted access in depth/height; evaluate reachability and stability constraints. (Fig.~\ref{fig:experiments}(c), Table~\ref{tab:grasp_results})

\smallskip

\subsection{Comparative Metrics}
We utilize the following metrics for quantitative analysis:

\begin{itemize}
    \item Grasp Accuracy \textbf{(GA)}: The Euclidean distance error $e_{trans} = \| \mathbf{p}_{pred} - \mathbf{p}_{gt} \|_2$ between the network's predicted 3D grasp center $\mathbf{p}_{pred} \in \mathbb{R}^3$ and the ideal ground-truth grasp center $\mathbf{p}_{gt}$, obtained from the state of the simulator object and from the offline annotation in hardware trials.
    
    \smallskip
    \item Search-to-Grasp Latency \textbf{(SGL)}: The total elapsed time from receiving the language query, executing the active search maneuvers, to finalizing the physical approach vector.
    \textit{(Note: SGL measures end-to-end mission time (distinct model inference latency))}
    \smallskip
    
    \item Collision-Free Grasp Rate \textbf{(CFGR)}: The number of times a complete end-to-end collision-free grasp is found and executed such that the object is lifted stably.
    
    \smallskip

    \item Collision-Induced Failures \textbf{(CIF)}: Number of trials where grasp execution fails due to collision in clutter.

    \item Object Search Failures \textbf{(OSF)}: The number of times the pipeline is unable to find the correct object.
    
\end{itemize}

\vspace{0.25em}

\subsection{Baseline Comparisons}
To the best of our knowledge, recent aerial manipulation frameworks were primarily developed for environments where the target object is clearly visible and isolated \cite{bauer2025open, bauer2023autonomous}. These methods lack explicit design considerations or evaluations for dense, cluttered scenes involving occlusions. Consequently, we evaluate our proposed method against representative baselines under an identical cluttered experimental setup to assess comparative robustness (see Table~\ref{tab:baseline}). 

Specifically, we compare against Bauer et al.~\cite{bauer2025open}, a centroid-based grasping approach, and Ramon-Soria et al. \cite{ramon2020grasp}, a traditional 6-DoF aerial manipulation baseline. Notably, while centroid-based methods such as [4] may exhibit a superficially lower Grasp Accuracy (GA) error by trivially minimizing the Euclidean distance to the object's geometric center, this geometric simplification fails to capture functional grasp regions of general objects. This limitation is evidenced by a severe drop in the Collision-Free Grasp Rate (CFGR) during our dense clutter trials. Furthermore, while \cite{ramon2020grasp} synthesizes 6-DoF grasp poses, its lack of active viewpoint adaptation and tightly coupled collision-feasibility evaluation results in high Object Search Failures (OSF) when subjected to severe occlusions.

\begin{table}[t]
\centering
\caption{Baseline Comparison Under Sparse and Dense Clutter (Scenario-1: Tabletop, Hardware Runs $N=30$)}
\label{tab:baseline}

\begin{threeparttable}

\resizebox{\columnwidth}{!}{%
\begin{tabular}{l|c|cccc}
\toprule
\textbf{Method} & \textbf{Clutter} & \textbf{GA (cm)}$\downarrow$ & \textbf{CFGR}$\uparrow$ & \textbf{OSF}$\downarrow$ & \textbf{CIF}$\downarrow$ \\
\midrule
Bauer-CoRL25~\cite{bauer2025open} & Sparse & 1.2 $\pm$ 0.4 & 43.33\% (13/30) & --\tnote{a} & 56.67\% (17/30) \\
RamonSoria-Eng20~\cite{ramon2020grasp} & Sparse & 4.6 $\pm$ 0.5 & 56.67\% (17/30) & 20\% (6/30) & 23.33\% (7/30) \\
\textbf{AeroGrab (Ours)} & Sparse & 2.2 $\pm$ 0.6 & \textbf{90\% (27/30)} & \textbf{3.33\% (1/30)} & \textbf{6.67\% (2/30)} \\
\midrule
Bauer-CoRL25~\cite{bauer2025open} & Dense & 2.4 $\pm$ 0.5 & 10\% (3/30) & --\tnote{a} & 90\% (27/30) \\
RamonSoria-Eng20~\cite{ramon2020grasp} & Dense & 5.5 $\pm$ 0.4 & 23.33\% (7/30) & 43.33\% (13/30) & 33.33\% (10/30) \\
\textbf{AeroGrab (Ours)} & Dense & 3.1 $\pm$ 0.6 & \textbf{80\% (24/30)} & \textbf{10\% (3/30)} & \textbf{10\% (3/30)} \\
\bottomrule
\end{tabular}}

\begin{tablenotes}
\footnotesize
\item[a] OSF not reported in Bauer-CoRL25; the target object is assumed to \\
        already be visible.
\end{tablenotes}

\end{threeparttable}
\end{table}

\begin{table}[t]
\centering
\caption{\footnotesize Grasping performance of the AeroGrab method across three scenarios.}
\label{tab:grasp_results}
\vspace{0.15cm}
\resizebox{\columnwidth}{!}{%
\begin{tabular}{lccccc}
\toprule
\multirow{2}{*}{Scenario} & \multicolumn{5}{c}{Metrics} \\
\cmidrule(lr){2-6}
& GA (cm) $\downarrow$ & SGL (s) $\downarrow$ & CFGR (\%) $\uparrow$ & OSF (\%) $\downarrow$ & CIF (\%) $\downarrow$ \\
\midrule
\multicolumn{6}{l}{\textit{Hardware Runs ($N=30$)}} \\
1 (Tabletop) & 3.1 $\pm$ 0.6 & 16.8 & 80.0 & 10.0 & 10.0 \\
2 (Window)   & 3.8 $\pm$ 0.6 & 10.5 & 73.33 & 10.0 & 16.67 \\
3 (Shelf)    & 4.9 $\pm$ 0.4 & 15.0 & 70.0 & 16.67 & 13.33 \\

\midrule
\multicolumn{6}{l}{\textit{Simulation Runs ($N=50$)}} \\
1 (Tabletop) & 1.2 $\pm$ 0.2 & 6.8 & 90.0 & 4.0 & 6.0 \\
2 (Window)   & 2.1 $\pm$ 0.4 & 4.7 & 86.0 & 6.0 & 8.0 \\
3 (Shelf)    & 1.9 $\pm$ 0.3 & 5.0 & 80.0 & 8.0 & 12.0 \\
\bottomrule
\end{tabular}}
\end{table}

\subsection{Ablation Study}
\label{sec:ablation}
To isolate the contributions of our algorithmic design choices, we ablate key components of the perception and planning stack. The evaluation is conducted on physical hardware in a standardized suite of cluttered tabletop trials $N={30}$. We evaluate against four distinct system configurations, focusing on the interplay between computational latency, environmental awareness, and physical execution success.

\begin{table}[b]
\centering
\caption{System Architecture Ablation Study (Hardware Runs, $N=30$)}
\label{tab:ablation}
\resizebox{\columnwidth}{!}{%
\begin{tabular}{l|*{5}{c}}
\toprule
\textbf{Architecture Variant} &
\textbf{Latency$^{\dagger}$ (ms)} $\downarrow$ &
\textbf{GA (cm)} $\downarrow$ &
\textbf{CFGR} $\uparrow$ &
\textbf{OSF} $\downarrow$ &
\textbf{CIF} $\downarrow$ \\
\midrule
No Obstacle Awareness & \textbf{24 $\pm$ 6} & 5.2 $\pm$ 0.3 & 33.33\% & 13.33\% & 53.34\% \\
No Real-Time Update   & 41 $\pm$ 10          & 4.5 $\pm$ 1.1 & 46.66\% & 33.34\% & 20.00\% \\
CPU-Only Collision    & 145 $\pm$ 12         & 3.8 $\pm$ 0.5 & 53.34\% & 23.33\% & 23.33\% \\
\midrule
\textbf{Full Pipeline (Ours)} & 41 $\pm$ 12 & \textbf{3.1 $\pm$ 0.6} & \textbf{80\%} & \textbf{10\%} & \textbf{10\%} \\
\bottomrule
\multicolumn{6}{l}{\scriptsize $^{\dagger}$Latency is the perception-to-plan compute time ( grasp scoring + feasibility evaluation) per planning cycle, excluding robot motion time.}
\end{tabular}%
}
\end{table}

\subsubsection{Experimental Protocol \& Configurations}
We compare our full proposed model against three degraded baselines:
\begin{itemize}
    \item \textbf{No Obstacle Awareness (Kinematic-Only):} Bypasses the collision check. Grasps are selected purely based on the CNN-ViT visual affordance score, ignoring the point cloud ${P}_{scene}$.
    \item \textbf{No Real-Time Update (Open-Loop):} The pipeline computes the optimal grasp and approach vector at $t=0$. The execution proceeds blindly without updating the robot dynamic hulls ${H}_{arm}(\mathbf{q})$ or tracking the target ${M}_{obj}$ during the approach phase.
    \item \textbf{CPU-Only Collision (Standard Pipeline):} Retains full obstacle awareness but replaces our GPU-accelerated batched hybrid scoring with a standard CPU-based spatial tree query (e.g., Octree or Bounding Volume Hierarchy) \cite{hornung2013octomap, pan2012fcl}. This mimics the standard collision-checking pipelines found in traditional ROS-based manipulation frameworks.
    \item \textbf{Full Pipeline (Ours):} The complete \textbf{AeroGrab} framework as detailed in Section III.
\end{itemize}

\subsection{Results and Discussion}

Table~\ref{tab:ablation} reports an ablation over three design choices: obstacle-aware feasibility reasoning, closed-loop perception updates, and low-latency collision evaluation. Removing obstacle awareness yields the lowest perception-to-plan latency (24\,ms) but severely degrades reliability, reducing CFGR to 33.33\% and increasing collision-induced failures (CIF) to 53.34\%. This indicates that many high-confidence grasps become infeasible once full-body clearance of the aerial manipulator and nearby clutter is enforced.

Disabling real-time perception updates (open-loop execution) increases grasp error to 4.5\,cm and reduces CFGR to 46.66\%, while also raising object search failures (OSF) to 33.34\%. This behavior is consistent with state drift and unmodeled disturbances during approach, where the system cannot correct target pose and clearance estimates online.

Replacing the GPU-based collision evaluation with a CPU-only pipeline increases latency to 145\,ms and achieves CFGR 53.34\%, but still incurs substantial OSF (23.33\%) and CIF (23.33\%). This suggests that slower feasibility updates can make the planner less responsive to rapidly changing geometry and state, leading to missed opportunities and late-stage failures. With all modules enabled, the full pipeline achieves the best overall performance, reaching CFGR 80\% with the lowest grasp error (3.1\,cm) and reduced failure rates (OSF 10\%, CIF 10\%) at a moderate compute latency of 41\,ms. Overall, reliable aerial grasping in clutter requires jointly combining collision-aware feasibility reasoning, fast feasibility updates, and closed-loop perception during the approach.

\section{Conclusion and Future Work}
We presented a language-driven aerial grasping pipeline that integrates target grounding, real-time 6-DoF grasp prediction, and platform-aware feasibility filtering for operation in cluttered environments. By enforcing kinematic and full-body collision constraints during grasp selection, the system generates executable grasp approaches while maintaining real-time perception-to-plan compute latency on lightweight onboard hardware. Hardware experiments achieve an 80\% collision-free grasp rate with a 41\,ms perception-to-plan latency and an average end-to-end episode time of 16.8\,s.

Future work will focus on deploying the complete pipeline fully onboard (including segmentation) and extending operation to GPS-denied environments without external infrastructure such as motion-capture systems, enabling autonomous aerial manipulation in real-world field conditions.

\section*{ACKNOWLEDGMENT}
The authors acknowledge the use of large language models (Gemini and ChatGPT) solely for refining the linguistic flow of this manuscript.

\bibliographystyle{IEEEtran}
\bibliography{final_iros}

\end{document}